\documentclass[sensors,article,accept,pdftex,moreauthors]{Definitions/mdpi} 

\usepackage{subscript}
\usepackage{amssymb}
\usepackage{afterpage}
\usepackage{graphicx}          
\usepackage{xcolor}
\usepackage[labelformat=simple]{subcaption}

\DeclareCaptionLabelFormat{subcaptionlabel}{\normalfont(\textbf{#2}\normalfont)}
\firstpage{1} 
\pubvolume{1}
\issuenum{1}
\articlenumber{0}
\pubyear{2024}
\copyrightyear{2024}
\externaleditor{Academic Editor: Konstantinos Sfikas}
\datereceived{21 October 2024} 
\daterevised{27 November 2024} 
\dateaccepted{10 December 2024} 
\datepublished{ } 
\hreflink{https://doi.org/} 

	\Title{IGAF: Incremental Guided Attention Fusion for Depth Super-Resolution}
	
	\TitleCitation{IGAF: Incremental Guided Attention Fusion for Depth Super-Resolution}
	
	\Author{Athanasios Tragakis $^{1}$, Chaitanya Kaul $^{2,}$*, Kevin J. Mitchell $^{1}$, Hang Dai $^{2}$, Roderick Murray-Smith $^{2}$\linebreak and Daniele Faccio $^{1}$\orcidA{}
		
	}
	
	\AuthorNames{Athanasios Tragakis, Chaitanya Kaul, Kevin J. Mitchell, Hang Dai, Roderick Murray-Smith and Daniele Faccio}
	
	\AuthorCitation{Tragakis, A.; Kaul, C.; Mitchell, K.J.; Dai, H.; Murray-Smith, R.; Faccio, D.}
	
	\address{$^{1}$ \quad School of Physics and Astronomy, University of Glasgow, Glasgow G12 8QQ, UK; a.tragakis.1@research.gla.ac.uk~(A.T.); kevin.mitchell@glasgow.ac.uk~(K.J.M.); daniele.faccio@glasgow.ac.uk~(D.F.)
		\\
		$^{2}$ \quad School of Computing Science, University of Glasgow, Glasgow G12 8QQ, UK; hang.dai@glasgow.ac.uk~(H.D.); roderick.murray-smith@glasgow.ac.uk~(R.M.-S.)\\
		
	}

	\corres{Correspondence: chaitanya.kaul@glasgow.ac.uk}
	
	\abstract{Accurate depth estimation is crucial for many fields, including robotics, navigation, and medical imaging. However, conventional depth sensors often produce low-resolution (LR) depth maps, making detailed scene perception challenging. To address this, enhancing LR depth maps to high-resolution (HR) ones has become essential, guided by HR-structured inputs like RGB or grayscale images. We propose a novel sensor fusion methodology for guided depth super-resolution (GDSR), a technique that combines LR depth maps with HR images to estimate detailed HR depth maps. Our key contribution is the Incremental guided attention fusion (IGAF) module, which effectively learns to fuse features from RGB images and LR depth maps, producing accurate HR depth maps. Using IGAF, we build a robust super-resolution model and evaluate it on multiple benchmark datasets. Our model achieves state-of-the-art results compared to all baseline models on the NYU v2 dataset for $\times 4$, $\times 8$, and $\times 16$ upsampling. It also outperforms all baselines in a zero-shot setting on the Middlebury, Lu, and RGB-D-D datasets. Code, environments, and models are available on GitHub.}

	
\begin{document}

		\section{Introduction}
		
		Accurate and useful visual perception is conventionally achieved by using RGB and depth sensors. Depth sensors, due to their small form factor, low cost, and low power consumption, are very popular in many fields of research such as robotics \cite{mav, Quadrotor, collison}, medical imaging \cite{cerebrovascular, endoscopy}, augmented reality and consumer electronics. However, they typically tend to have lower spatial resolution than conventional imaging modalities such as RGB, leading to information loss, which can be overcome with accurate super-resolution techniques. In order to achieve this high resolution, existing techniques leverage correlations between sharp high-frequency texture edges of RGB images and low-resolution edge discontinuities of depth images. Typical super-resolution solutions often prove inadequate for addressing depth super-resolution due to their limited ability to effectively incorporate the unique characteristics and complexities inherent in the depth data \cite{gdsr_survey}. To this end, the task of DSR is to provide solutions aimed at optimizing the super-resolution of lower resolution depth~maps. 
		
		The depth super-resolution literature can be broadly categorized into three different approaches: filtering, optimization, and learning-driven strategies. Filtering-driven\linebreak DSR~\cite{yang2007spatial, riemens2009multistep, liu2013joint, lo2017edge, sun2019weighted, qiao2021fast} relies on finding filters based on neighboring pixels in the LR depth map and the HR guidance image. A downside of this approach is the creation of artifacts and errors when the scenes that need to be super-resolved are complex. Optimization-driven DSR~\cite{optim1, optim2, optim3, optim4} converts the task of super-resolution to an optimization problem where a cost function between the LR depth map and the HR depth map is minimized. This approach relies on selecting an appropriate cost function and is highly sensitive to the choice made. Finally, learning-driven approaches~\cite{riegler2016atgv, learning1, learning2, channelattention} in recent years have made use of deep learning techniques which have quickly become the de facto preferred solution of choice in the field of DSR.

		In GDSR, most approaches rely on fusion techniques after the feature extraction stage. The unique per-modality features are fused together to create an HR depth map. It is important to have both strong feature extractors and fusion modules for this task as inadequate feature extractors do not provide the appropriate unique information on crucial features that help create sharp reconstructions.
		The RGB image is responsible for providing structure so that the end result does not suffer from depth bleeding. This task is not trivial, as over-transfer may occur and non-relevant structures from the RGB image can be transferred to the depth map. An example would be an image on the cover of a book where the network should be able to identify the textures that are irrelevant to depth reconstruction. Equally important to the feature extractors are the fusion modules that combine the two branches and refine the available information to estimate clear depths.

\begin{figure}[H]
				\includegraphics[trim={11.1cm 10.5cm 11.5cm 5.7cm}, scale=2, clip=true]{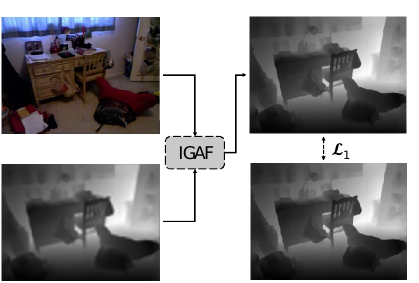}
				\caption{Overview of the proposed multi-modal architecture for the guided depth super resolution estimation.} 
				\label{overview}
		\end{figure}

		Existing works face two major limitations: (1) weak feature extractors that fail to capture the distinct and complementary characteristics of RGB and depth modalities, and (2) naive fusion strategies, such as simple concatenation or addition, which result in modality-specific artifacts, including over-transfer of irrelevant RGB features and insufficient depth-specific enhancement. These issues often lead to blurring, depth bleeding, and misaligned structures in the final depth maps. Our approach (see Figure \ref{overview} for an overview), IGAF, systematically addresses these challenges (the {code} is available on \url{https://github.com/Thanos-DB/IncrementalGuidedAttentionFusion} accessed on 9 December 2024). 
        First, it introduces the filtered wide-focus (FWF) block to extract per-modality features with greater sensitivity to spatial textures and channel importance. Second, it employs an Incremental guided attention fusion (IGAF) module that refines features iteratively. Unlike previous methods, which fuse features in a single stage, IGAF leverages cross-modal attention to ensure only relevant information is transferred while suppressing noise. By iteratively fusing and refining features, our method minimizes artifacts, resulting in sharper, more accurate depth maps. To formalize, our contributions are as follows:
		\begin{itemize}
			\item We propose the incremental guided attention fusion $\mathbf{(IGAF)}$ model, which surpasses existing works for the task of DSR on all tested benchmark datasets for all tested benchmark resolutions.
			\item We propose the $\mathbf{IGAF}$ module, which is a flexible and adaptive attention fusion strategy, with the ability to effectively fuse multi-modal features by creating weights from both modalities and then applying a two-step cross fusion.
			\item We propose the filtered wide-focus $\mathbf{(FWF)}$ block, a strong feature extractor composed of two modules, the feature extractor $\mathbf{(FE)}$ and wide-focus $\mathbf{(WF)}$. The $\mathbf{FE}$ with the help of channel attention is able to highlight relevant feature channels in feature volumes, and the $\mathbf{WF}$ using varying dilation rates in the convolution layers per branch, creates multi-receptive field spatial information that allows integrating global resolution features using dynamic receptive fields to better highlight textures and edges. The combination of the two forms a general-purpose feature extractor specifically tailored towards DSR. 
		\end{itemize}

				\vspace{-11pt}

		\section{Literature Review}

		$\textbf{Depth Super-Resolution Architectures.}$ DSR techniques are broadly categorized into those that use RGB or grayscale images as guidance, and those that do not. 
		Non-guided DSR techniques \cite{riegler2016atgv, ye2020depth, huang2019pyramid} try to solve the task by only using an LR depth map. This results in a simplified data acquisition pipeline (as syncing different modalities is not required, leading to smaller-sized datasets) alongside a simpler model by alleviating the need for sensor fusion techniques for the additional stream. This simplicity in the data processing, however, comes at the cost of producing smooth edges, especially on the contours of objects, as well as blurring and distortion effects in the super-resolved depth maps. 
		
		GDSR techniques propose a fix to over-smoothed edges by using structural and textural information from RGB or grayscale images. Additional techniques are needed to prevent the over-transfer of information from the guidance stream and to only retain the features that are relevant. Ref. \cite{fdsr} propose a fast model utilizing the high-frequency information of the guidance RGB stream using octave convolutions, but fuse the information from the two branches by a simple concatenation. Ref. \cite{suft}, on the other hand, propose to fuse information between the two modalities through a symmetric uncertainty incorporated into their system. Ref. \cite{jiif} use a joint implicit function representation to learn the interpolation weights and values for the HR depth simultaneously. Ref. \cite{ctkt} employ knowledge distillation such that the guidance stream is only needed during training while simplifying the model during the test phase. Ref. \cite{bridgenet} utilize bridges to fuse information during multi-task learning. The two tasks in their system are depth super-resolution and monocular depth estimation. Additional novel techniques include \cite{DCTNet, Spherical, ctkt}. 
		
		$\textbf{Attention Feature Fusion in Depth Super-Resolution.}$ Feature fusion techniques are crucial for multi-modal data processing. They range from a simple addition or concatenation of multiple features to complex feature processing modules. Ref. ~\cite{multiscaleattentionfusion} place a hierarchical attention fusion module in the generator of a generative adversarial network towards this task. Ref. ~\cite{AMBFF} employ an attention fusion strategy to adaptively utilize information from both modalities by first enhancing the features and then using an attention mechanism to fuse the two branches. Ref. ~\cite{WAFP-Net} also propose a two-step approach where first a weighted attention fusion, followed by a high-frequency reconstruction generates the resulting high-resolution depth image. Ref. ~\cite{channelattention} use channel attention combined with reconstruction in the proposed module, whilst \cite{MMAF} have a fusion module consisting of feature enhancement and feature re-calibration step. 
		
\begin{figure}[H]
				\includegraphics[trim={6cm 10cm 4.5cm 6cm}, clip=true, scale=.72]{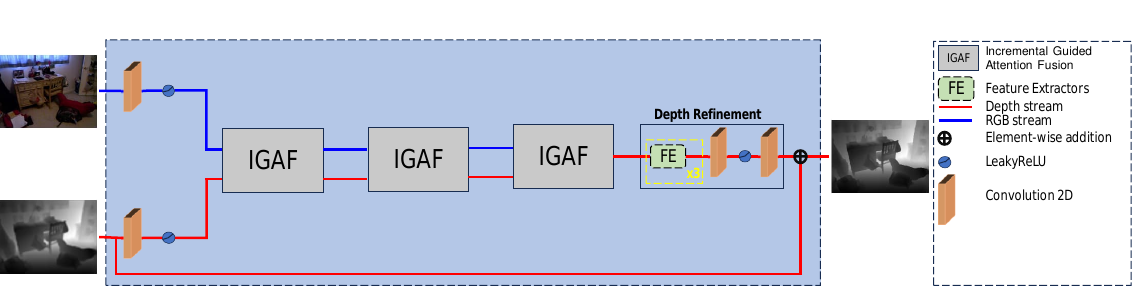}
				\caption{The proposed multi-modal architecture utilizes information from both an LR depth map and an HR RGB image. Firstly, each modality passes through a convolutional layer followed by a LeakyReLU activation. The model utilizes the IGAF modules to combine information from the two modalities by fusing the relevant information on each stream and ignoring information that is unrelated to the depth maps. Finally, after the third IGAF module, the depth maps are refined and added using a global skip connection from the original upsampled LR depth maps. The RGB modality is used to provide guidance to estimate an HR depth map given an LR one.}
				\label{overview2}
			\end{figure}

		Existing works fail to effectively leverage both modalities to create fusion weights that accurately propagate relevant features. This limitation often results in the over-transfer of RGB features or insufficient depth-specific enhancements, leading to artifacts such as depth bleeding and texture misalignment. Our approach (see Figure \ref{overview2}) overcomes these drawbacks through a flexible and more powerful attention-based mechanism. By creating weights from one modality to iteratively guide the fusion with the other, we ensure that only the most relevant features are propagated. Unlike previous methods that rely on simple concatenation or addition for fusion, we introduce an Incremental guided attention fusion (IGAF) module that performs cross-modal attention in iterative steps. This process eliminates the over-transfer of RGB features while emphasizing critical depth-specific information. Specifically, we carry this out by first creating a naive fusion of the RGB and depth modalities (element-wise multiplication), followed by creating a structural guidance for the depth modality by learning a set of attention weights from the naive fusion for the RGB image (the first spatial attention fusion, i.e., SAF block; see Figure below). We then use this intermediate fusion as structural guidance for the depth image to create a better fusion output than existing methods (the second SAF block).

		\section{Methodology}
		\subsection{Problem Statement}
		Consider a dataset $\{\mathbf{G},\mathbf{L},\mathbf{H}\}$, where $\mathbf{G}$ represents the RGB images, $\mathbf{L}$ represents the RGB images' corresponding LR depth maps and $\mathbf{H}$ are the corresponding HR depth maps, for each RGB image $\mathbf{g}_{i} \in \mathbb{R}^{3 \times sH \times sW}$. Each LR depth map is $\mathbf{l}_{i} \in \mathbb{R}^{1 \times H \times W}$ and HR depth map is $\mathbf{h}_{i} \in \mathbb{R}^{1 \times sH \times sW}$, where H and W are the spatial resolutions of the images. The 1 in $\mathbf{l}_{i}$ (and $\mathbf{h}_{i}$) and 3 in $\mathbf{g}_{i}$ refer to the number of input channels, while s is the scale factor between the HR and the LR depth maps. The model estimates an HR depth map $\hat{\mathbf{H}}$ where $\hat{\mathbf{h}}_i \in \mathbb{R}^{1 \times sH \times sW}$, by first upsampling $\mathbf{L}$ to $\mathbf{L}_{U}$ where $\mathbf{l}_{Ui} \in \mathbb{R}^{1 \times sH \times sW}$ using bicubic interpolation such that the dimensions between $\mathbf{G}$ and $\mathbf{L}_{U}$ match. A formal representation is:
		\begin{equation} \label{eq:1}
			\hat{\mathbf{H}} = \mathbf{L}_{U} + \mathcal{F}(\mathbf{G}, \mathbf{L}_{U}; \theta),
		\end{equation}
		\noindent where $\mathcal{F}(\cdot)$ is the learned function that maps $\mathbf{L}_{U}$ and $\mathbf{G}$ to $\hat{\mathbf{H}}$ for the predicted HR depth map. Finally, $\theta$ represents the learned parameters. The addition operation in Equation (\ref{eq:1}) represents the global residual connection as seen in Figure \ref{overview2}.
		
		\subsection{Model Architecture}
		We follow the conventional architecture of a dual-stream model as depicted in Figure \ref{overview2}. Our model contains two inputs, one for the RGB guidance and the other for the upsampled LR depth map. First, each modality is processed via a convolutional layer followed by a LeakyReLU activation. This is followed by 3 $\mathbf{IGAF}$ modules which extract and fuse the multi-modal features from the two input modalities. After the fusion modules, the depth is refined through our refinement block and a global skip connection adds the LR upsampled depth map to the final feature representation to produce the final prediction. The predicted depth map $\hat{\mathbf{H}}$ is calculated as \begin{equation}
			\hat{\mathbf{H}} = \mathbf{L}_{U} + \mathrm{Depth\_Refinement}(\mathrm{IGAF}(\mathrm{IGAF}(\mathrm{IGAF}(\alpha(\mathrm{Conv}(\mathbf{G})), \alpha(\mathrm{Conv}(\mathbf{L}_{U})))))),
		\end{equation}
		
		\noindent where $\alpha$ is the LeakyReLU activation. The depth refinement consists of 3 feature extractor modules (see Section \ref{igaf_module}), and a convolution-LeakyReLU-convolution stack of layers.

		\subsubsection{The $\mathbf{IGAF}$ Module}
		\label{igaf_module}
\begin{figure}[H]
				\includegraphics[trim={8.5cm 7.5cm 5.5cm 7cm}, clip=true, scale=.88]{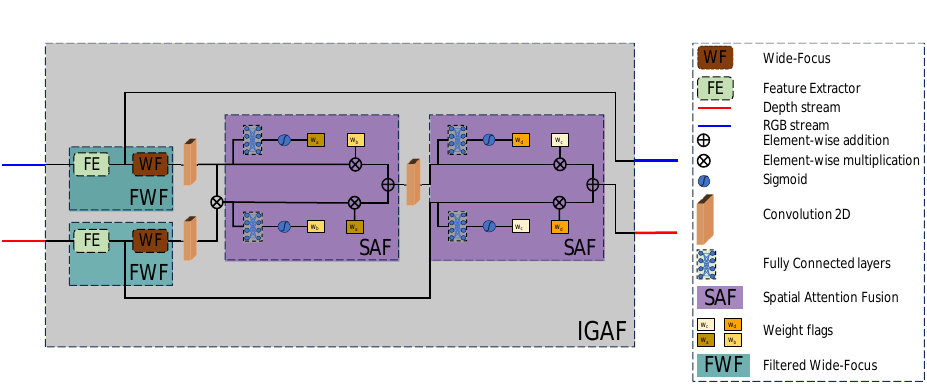}
				\caption{{The} $\mathbf{IGAF}$ module. The module is responsible for both feature extraction and modality fusion. Each modality passes through a feature extraction stage $\mathbf{(FWF)}$ before the initial naive fusion by an element-wise multiplication. An $\mathbf{SAF}$ block follows, which fuses the result of the multiplication with the extracted features of the RGB stream creating an initial structural guidance. The second $\mathbf{SAF}$ block incrementally fuses this extracted structural guidance with the depth stream. The output of each $\mathbf{SAF}$ block is generated by learning attention weights and subsequently performing a cross-multiplication operation between the two input sequences, resulting in fused and salient processed information.}
				\label{fig:igaf_module}
			\end{figure}

		Each $\mathbf{IGAF}$ module processes two inputs and provides two outputs (Figure \ref{fig:igaf_module}). For the last $\mathbf{IGAF}$ module, we only propagate the depth stream forward into the depth refinement block and ignore the second output.
		
				\vspace{-11pt}

\begin{figure}[H]
				\includegraphics[trim={5.7cm 7.5cm 5.5cm 7.4cm}, scale=.77, clip=true]{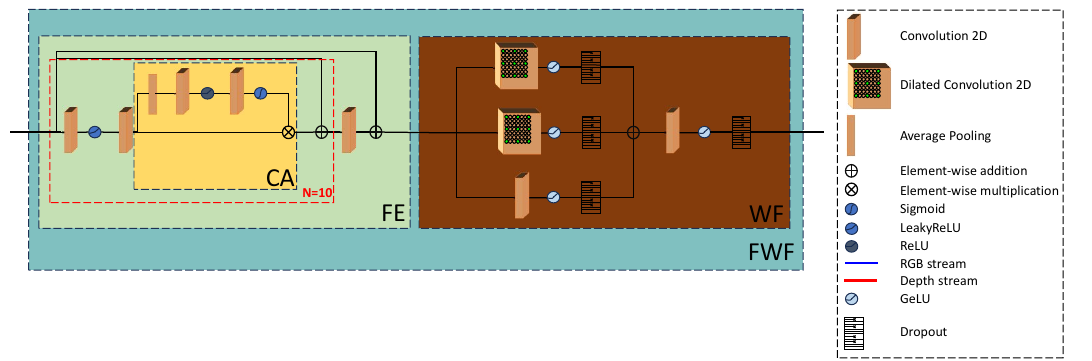}
				\caption{Overview of the $\mathbf{FWF}$ module. The two modules are separated and not combined into one larger module because the propagation of shallower features through the skip connections as seen in Figure \ref{fig:igaf_module} boosts the performance of the model. The $\mathbf{FE}$ module is a series of convolutional layers, a channel attention process, and two skip connections. The $\mathbf{WF}$ module uses linearly increasing dilation rates in convolutional layers to extract multi-resolution features.}
				\label{fe_wf}
			\end{figure}

		At first, each stream passes through a two-piece feature extraction block, the $\mathbf{FWF}$ consisting of a general feature extractor ($\mathbf{FE}$) and a $\mathbf{WF}$ block. The $\mathbf{FE}$ processes the input using a convolution--LeakyReLU--convolution stack of layers. Next, a channel attention $\mathbf{CA}$ module focuses only on the relevant channels while reducing the influence of the less important or noisy ones. Finally, we employ an element-wise addition between the input of the module and the output of the $\mathbf{CA}$ module, followed by another convolutional layer, and a skip connection that is global within the $\mathbf{FE}$ module which propagates the global structure of the depth forward through the model. For simplicity in the explanations and equations, we treat N in the Figure below as 1, although during training N = 10 was used. The choice of the $\mathbf{CA}$ module was empirically estimated after multiple trainings, alternating between channel attention, spatial attention, and a combination of both. For the $\mathbf{CA}$ module, we have:
		
		\begin{equation}
			\mathbf{F}_{CA} = \mathbf{K} \times \sigma ( \mathrm{Conv(ReLU(Conv(Avg.Pool(\mathbf{K})))))},
		\end{equation}
		\noindent where $\mathbf{F}_{CA}$ represents the feature maps output by the $\mathbf{CA}$ module, $\mathbf{K}$ is the feature maps input and $\sigma$ is the sigmoid activation. The $\mathbf{FE}$ module is represented as:
		\begin{equation}
			\mathbf{F}_{FE} = \mathbf{M} + \mathrm{Conv(\mathbf{M}+CA(Conv(} \alpha \mathrm{(Conv(\mathbf{M})))))},
		\end{equation}
		\noindent where $\mathbf{F}_{FE}$ is the feature maps output by the $\mathbf{FE}$ module, $\mathbf{M}$ is the feature maps input and $\alpha$ is the LeakyReLU activation. See Figure \ref{fe_wf}.
		
		$\mathbf{WF}$ is an efficient feature extractor first introduced by \cite{fct} and\cite{tragakis2024glfnet} 
        for medical image segmentation and has shown great promise in extracting multi-scale features from feature representations. It contains three branches, each with a different dilation rate for the convolution kernels, followed by an activation layer and a dropout layer to prevent overfitting. After the element-wise addition, another convolutional layer extracts features from the gradually increased receptive fields of the $\mathbf{WF}$ dilated convolution layer to aggregate the extracted multi-resolution features. This is followed by an activation layer and a dropout layer to avoid overfitting. A $\mathbf{WF}$ block is used after every $\mathbf{FE}$ module to aggregate the multi-resolution hierarchical features extracted in each layer. The RGB stream after the $\mathbf{FE}$ module uses a skip connection to propagate the extracted features to the next $\mathbf{IGAF}$ module further propagating the global scene structure forward within the model. We observed from our experiments that not placing a skip connection after $\mathbf{WF}$ or at later stages is an effective strategy for learning better scene structure as the forward propagation of shallower features without the skip connection helps propagate high-frequency structure better through the model, which can be verified through our ablations in.
		
		The first fusion in the $\mathbf{IGAF}$ module (see Figure \ref{fig:igaf_module}) is an element-wise multiplication of both modalities. Its result (a) creates intermediate feature weights $\mathbf{w}_{b}$ and (b) is used in an element-wise weighted addition between the two intermediate features of the $\mathbf{IGAF}$ block. Similarly, the extracted features from the RGB stream (a) create intermediate feature weights $\mathbf{w}_{a}$ and (b) are the second component of the weighted addition. The addition can be seen as adding features from a joint representation of the RGB and LR images weighted by their common features. This helps the model focus on both high-level semantic structures in the image through the depth features, as well as high-frequency features from the RGB images while weighting them according to backpropagation. For each component, weights are extracted and applied in a crosswise fashion, i.e., weights from one component are applied to the other component, resulting in a spatial attention fusion $\mathbf{(SAF)}$ block. This allows the model to learn features across both modalities that can limit their influence on the output resulting in a smoother depth map at the output. The weights are learnable and created via two-layer MLPs. A formalized expression of the SAF block is:
		\begin{equation}
			\mathrm{y}_{A} = (\mathrm{x}_{A} \mathrm{A}^{T}_{A1} + \mathrm{b}_{A1}) \mathrm{A}^{T}_{A2} + \mathrm{b}_{A2} 
			\label{first}
		\end{equation}
		\begin{equation}
			\mathrm{y}_{B} = (\mathrm{x}_{B} \mathrm{A}^{T}_{B1} + \mathrm{b}_{B1}) \mathrm{A}^{T}_{B2} + \mathrm{b}_{B2}
			\label{second}
		\end{equation}
		\begin{equation}
			\mathrm{y} = \mathrm{x}_{A} \sigma (\mathrm{y}_{B}) + \mathrm{x}_{B} \sigma (\mathrm{y}_{A}),
			\label{final}
		\end{equation}
		\noindent where $\mathrm{x}_{A}$ and $\mathrm{x}_{B}$ are the two inputs of the block, $\mathrm{A}$ represents the weights, $\mathrm{b}$ the bias term and $\sigma$ the sigmoid activation. Equations (\ref{first}) and (\ref{second}) show the 2 MLP layers for the weights creation that are used cross-wise with the inputs as seen in Equation (\ref{final}).
		
		The output of the first $\mathbf{SAF}$ block passes through a convolutional layer for joint feature processing and is then used as input for the second $\mathbf{SAF}$ block. This convolution layer now extracts shared features from the two fused modalities. The second $\mathbf{SAF}$ block works in a similar manner to the previous one but now fuses the joint features from the two modalities with the depth features. The first $\mathbf{SAF}$ module fuses together the extracted features from the RGB stream and the naive feature fusion obtained by the element-wise multiplication. The second $\mathbf{SAF}$ block fuses together the results of the first $\mathbf{SAF}$ block after the convolutional layer and the output of the $\mathbf{FE}$ module of the depth stream. This fusion is incremental in nature as we iteratively combine RGB and depth features in multiple steps to create a cross-modal fusion of attributes leading to simultaneously processing both structure and depth.
		
		\section{Experiments}
		
		We test our model on four benchmark datasets commonly used widely in comparing proposed models for the DSR task. These are the NYU v2 \cite{nyuv2}, Middlebury \cite{middlebury_1, middlebury_2}, Lu \cite{lu} and RGB-D-D \cite{fdsr} datasets.
		
		We only train $\mathbf{IGAF}$ on the NYU v2 dataset and do not fine-tune the model further on the others. Our results on the remaining datasets are a zero-shot prediction to demonstrate the generalization ability of our model. The NYU v2 contains 1449 pairs of RGB and depth images. The first 1000 images are used to train the model and the remaining 449 are used for the purpose of evaluating our approach. For the Middlebury dataset, we use the provided 30 RGB and depth image pairs and for the Lu dataset we use the 6 pairs following previous works \cite{lu, suft, jiif, fdsr} in order to report consistent results with other proposed works. For RGB-D-D, we use 405 RGB and depth image pairs following \cite{fdsr}.
		
		$\textbf{Implementation Details:}$ We run all our experiments using one RTX 3090 GPU using PyTorch 2.0.1. The initial learning rate for our model is set to 0.00025 and is reduced by half using the MultiStepLR scheduler with milestones. The milestones are set to every 25 epochs with the exception of the last one being 150 out of a total of 200 epochs. A batch size of 1 is used to train the model. We use the Adam optimizer and the Root Mean Square Error (RMSE) as our metric to report all our results.
		We use the $\mathcal{L}_{1}$ loss to train our model:
		\begin{equation}
			\mathcal{L}_{1} = \frac{1}{N} \sum_{i=1}^{n} |h_i - \hat{h}_i|,
		\end{equation}
		\noindent where $N$ is the number of pixels, $h_i$ is the ground truth depth map and $\hat{h}_i$ is the predicted depth map.
		
		During training, we use $256 \times 256$ patches of the HR image that are randomly cropped. The LR depth maps are simulated by bicubic downsampling which is consistent with other approaches using the same datasets. Additionally, we evaluate our model using the ``real-world manner'' RGB-D-D dataset where both HR and LR are provided. The dimensions of the LR depth map is $192\times144$ and the HR depth map is $512\times384$.
		
		\section{Results}
		
\begin{figure}[H]
			\centering
			
			\begin{subfigure}{0.25\textwidth} 
				\includegraphics[width=\linewidth]{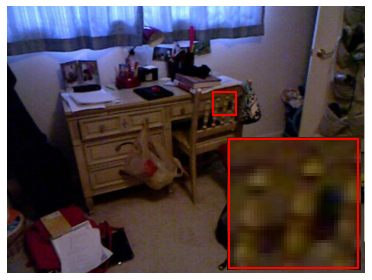}
			\end{subfigure}%
			\begin{subfigure}{0.25\textwidth}
				\includegraphics[width=\linewidth]{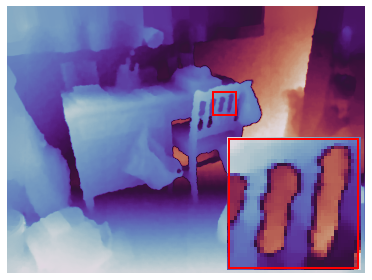}
			\end{subfigure}%
			\begin{subfigure}{0.25\textwidth}
				\includegraphics[width=\linewidth]{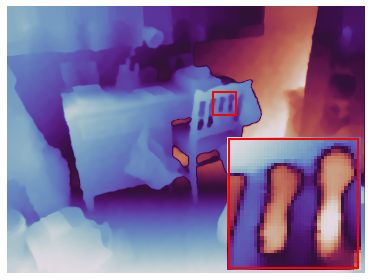}
			\end{subfigure}%
			\begin{subfigure}{0.25\textwidth}
				\includegraphics[width=\linewidth]{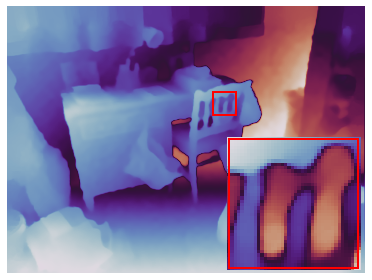}
			\end{subfigure}%
			
			\begin{subfigure}{0.25\textwidth}
				\includegraphics[width=\linewidth]{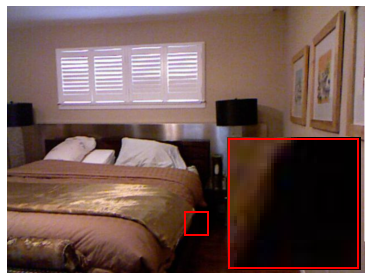}
			\end{subfigure}%
			\begin{subfigure}{0.25\textwidth}
				\includegraphics[width=\linewidth]{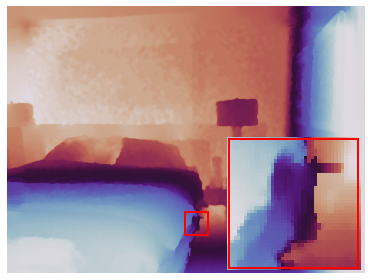}
			\end{subfigure}%
			\begin{subfigure}{0.25\textwidth}
				\includegraphics[width=\linewidth]{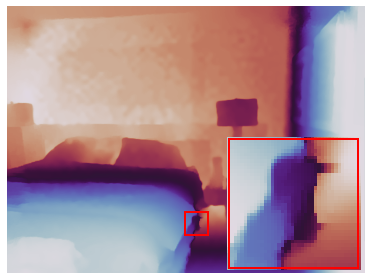}
			\end{subfigure}%
			\begin{subfigure}{0.25\textwidth}
				\includegraphics[width=\linewidth]{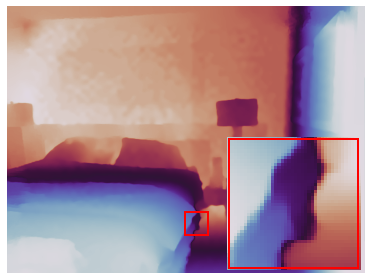}
			\end{subfigure}%
			
			\begin{subfigure}{0.25\textwidth}
				\includegraphics[width=\linewidth]{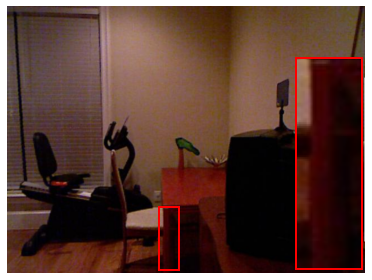}
			\end{subfigure}%
			\begin{subfigure}{0.25\textwidth}
				\includegraphics[width=\linewidth]{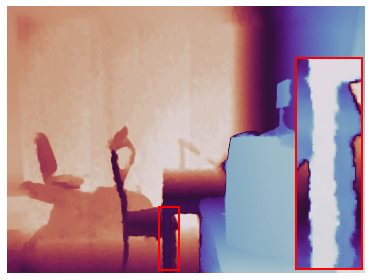}
			\end{subfigure}%
			\begin{subfigure}{0.25\textwidth}
				\includegraphics[width=\linewidth]{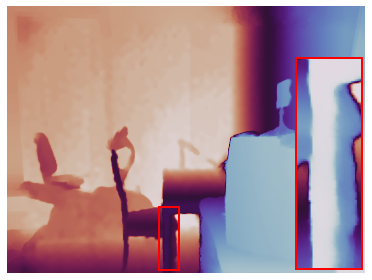}
			\end{subfigure}%
			\begin{subfigure}{0.25\textwidth}
				\includegraphics[width=\linewidth]{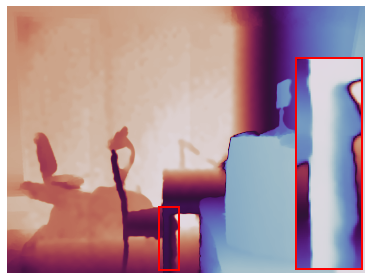}
			\end{subfigure}%

			\begin{subfigure}{0.25\textwidth}
				\includegraphics[width=\linewidth]{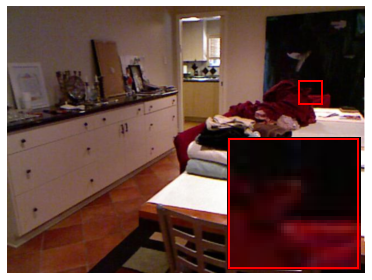}
				\captionsetup{justification=centering} 
				\caption{RGB}
			\end{subfigure}%
			\begin{subfigure}{0.25\textwidth}
				\includegraphics[width=\linewidth]{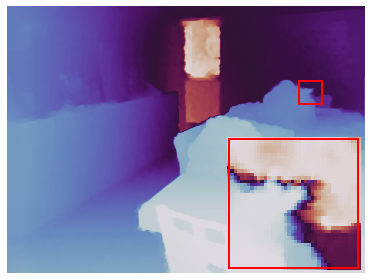}
				\captionsetup{justification=centering} 
				\caption{GT}
			\end{subfigure}%
			\begin{subfigure}{0.25\textwidth}
				\includegraphics[width=\linewidth]{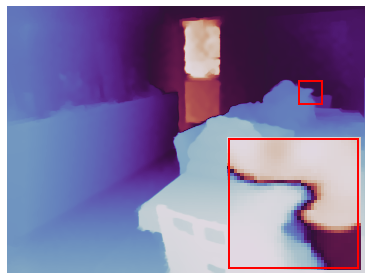}
				\captionsetup{justification=centering} 
				\caption{IGAF}
			\end{subfigure}%
			\begin{subfigure}{0.25\textwidth}
				\includegraphics[width=\linewidth]{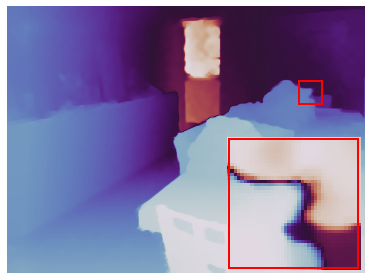}
				\captionsetup{justification=centering} 
				\caption{SUFT}
			\end{subfigure}%
			\caption{Qualitative comparison between our model and SUFT \cite{suft}. The visualizations shown are for the $\times 8$ case. Our model creates more complete depth maps as seen in (c) for rows 1 and 2. In (c), row 3 shows that our model creates sharper edges with minimal bleeding. Also, in (c), row 4 the proposed model creates less smoothing with less bleeding. (Colormap chosen for better visualization. Better seen in full-screen, with zoom-in options).}
			\label{qualitative_comparison}
		\end{figure}

		Our model achieves state-of-the-art (SOTA) results on all benchmark test datasets compared to the baselines demonstrating its ability to super-resolve various depth resolutions as well as its generalization capabilities across multiple datasets. The Tables below 
        show a quantitative comparison between our model and previous works. The evaluation is carried out based on the RMSE metric. The best performance is marked in bold. The Figure below showcases a qualitative comparison, on the NYU v2 dataset, between our model and SUFT \cite{suft}. The visualizations show how our proposed attention fusion helps alleviate problems such as bleeding and blurring that occur in previous SOTA models. This happens because IGAF iteratively refines features by leveraging structural guidance from RGB and selectively emphasizing depth-specific details. This approach minimizes the over-transfer of irrelevant RGB features, which is a key cause of blurring and bleeding in prior methods.
		The SAF blocks incrementally learn attention weights, ensuring sharper edges and reducing distortions. Qualitative comparisons in the Figure below demonstrate these advantages. See Figure~\ref{qualitative_comparison}.
		
		\vspace{-\baselineskip}
		
		\begin{table}[H]
			\caption{Results on the NYU v2 data set.}
				\begin{tabularx}{\textwidth}{CCCCCCCCCC}
					\toprule
					{\bf Method} & {\bf Bicubic} & {\bf DG}  & {\bf SVLRM}  & {\bf DKN}  & {\bf FDSR}  & {\bf SUFT} & {\bf CTKT}  & {\bf JIIF}   & {\bf IGAF}\\
					& & {\bf \cite{dg}} & {\bf \cite{svlrm}} & {\bf \cite{dkn}} & {\bf \cite{fdsr}}  & {\bf \cite{suft}} & {\bf \cite{ctkt}} & {\bf \cite{jiif}} & \\
					\midrule
					$\times 4$ & 8.16 & 1.56 & 1.74 & 1.62 & 1.61 & 1.14 & 1.49 & 1.37  & $\mathbf{1.12}$
					\\
					$\times 8$ & 14.22 & 2.99 & 5.59 & 3.26 & 3.18 & 2.57 & 2.73 & 2.76  & $\mathbf{2.48}$  \\
					$\times 16$ & 22.32 & 5.24 & 7.23 & 6.51 & 5.86 & 5.08 & 5.11 & 5.27  & $\mathbf{5.00}$\\\bottomrule
				\end{tabularx}
			\label{nyuv2}
		\end{table}
		
		\vspace{-\baselineskip}
		
		\begin{table}[H]
			\caption{Results on the RGB-D-D data set.}
				\begin{tabularx}{\textwidth}{CCCCCCCCCC}
					\toprule
					{\bf Method} & {\bf Bicubic} & {\bf DJFR}  & {\bf PAC}  & {\bf DKN} & {\bf FDKN} & {\bf FDSR}  & {\bf JIIF}  & {\bf SUFT} & {\bf IGAF}\\
					& & {\bf \cite{djfr}} & {\bf \cite{pac}} & {\bf \cite{dkn}} & {\bf \cite{dkn}} & {\bf \cite{fdsr}} & {\bf \cite{jiif}} & {\bf \cite{suft}} & \\
					\midrule
					$\times 4$ & 2.00 & 3.35 & 1.25 & 1.30 & 1.18 & 1.16 & 1.17 & 1.20 & $\mathbf{1.08}$\\
					$\times 8$ & 3.23 & 5.57 & 1.98   & 1.96  & 1.91 & 1.82 & 1.79 & 1.77 & $\mathbf{1.69}$\\
					$\times 16$ & 5.16 & 7.99 & 3.49 & 3.42 & 3.41 & 3.06 & 2.87 & 2.81 & $\mathbf{2.69}$\\\bottomrule
				\end{tabularx}
			\label{rgbdd}
		\end{table}
		
		\vspace{-\baselineskip}
		
		\begin{table}[H]
			\caption{Results on the Lu data set.}
				\begin{tabularx}{\textwidth}{CCCCCCCCCC}
					\toprule
					{\bf Method} & {\bf Bicubic} & {\bf DMSG}  & {\bf DG}  & {\bf DJF}  & {\bf DJFR}  & {\bf PAC}    & {\bf JIIF} & {\bf DKN} & {\bf IGAF}\\
					&   &  {\bf \cite{dmsg}} & {\bf \cite{dg}} & {\bf \cite{djf}} & {\bf \cite{djfr}} & {\bf \cite{pac}} & {\bf \cite{jiif}} & {\bf \cite{dkn}} & \\
					\midrule
					$\times 4$ & 2.42 & 2.30 &2.06 & 1.65 & 1.15 & 1.20 & 0.85 & 0.96 & $\mathbf{0.82}$\\
					$\times 8$ &  4.54 & 4.17 &  4.19 & 3.96 & 3.57 & 2.33 & 1.73 & 2.16 & $\mathbf{1.68}$\\
					$\times 16$ & 7.38 & 7.22 & 6.90 &  6.75 & 6.77 & 5.19 & 4.16 & 5.11 & $\mathbf{4.14}$\\\bottomrule
				\end{tabularx}
			\label{lu}
		\end{table}
		
		\vspace{-\baselineskip}
		
		\begin{table}[H]
			\caption{Results on the ``real-world manner'' RGB-D-D data set.}
			\label{realworldmannerrgbdd}
				\begin{tabularx}{\textwidth}{cCCCCCCCCC}
					\toprule
					{\bf Method} & {\bf Bicubic} & {\bf DJF}  & {\bf DJFR}  & {\bf FDKN} & {\bf DKN} & {\bf FDSR}  & {\bf JIIF}  & {\bf SUFT} & {\bf IGAF}\\
					& & {\bf \cite{djf}} & {\bf \cite{djfr}} & {\bf \cite{dkn}} & {\bf \cite{dkn}} & {\bf \cite{fdsr}} & {\bf \cite{jiif}} & {\bf \cite{suft}} & \\
					\midrule
					``real-world manner'' & 9.15 & 7.90 & 8.01 & 7.50 & 7.38 & 7.50 & 8.41 & 7.17 & $\mathbf{7.01}$\\\bottomrule
				\end{tabularx}
			
		\end{table}
		
		\vspace{-\baselineskip}
		
		\begin{table}[H]
			\caption{Results on the Middlebury data set.}
			\label{middlebury}
				\begin{tabularx}{\textwidth}{cCCCCCCCCC}
					\toprule
					{\bf Method} & {\bf Bicubic} & {\bf PAC} & {\bf DKN} & {\bf FDKN}  & {\bf CUNet} & {\bf JIIF} & {\bf SUFT} & {\bf FDSR} & {\bf IGAF}\\
					& & {\bf \cite{pac}} & {\bf \cite{dkn}} & {\bf \cite{dkn}} & {\bf \cite{cunet}} & {\bf \cite{jiif}} & {\bf \cite{suft}} & {\bf \cite{fdsr}} & \\
					\midrule
					$\times 4$ & 2.28 & 1.32 & 1.23 & 1.08 & 1.10 & 1.09 & 1.20 & 1.13 & $\mathbf{1.01}$ \\
					$\times 8$ & 3.98 & 2.62 & 2.12 & 2.17 & 2.17 & 1.82 & 1.76 & 2.08 & $\mathbf{1.73}$ \\
					$\times 16$ & 6.37 & 4.58 & 4.24 & 4.50 & 4.33 & 3.31 & 3.29 & 4.39 & $\mathbf{3.24}$ \\\bottomrule
				\end{tabularx}
			
		\end{table}
		
		\vspace{-\baselineskip}

		\section{Ablation Study}
		We run ablations on NYU v2 for the $\times 4$ DSR scenario. We study the effects of addition and concatenation as a fusion strategy by replacing $\mathbf{IGAF}$ in our model with the two naive approaches. In Table \ref{nyuv2ablation1}, we show that our carefully built module based on empirical results outperforms them both as expected.
		
		\begin{table}[H]
			\caption{Demonstrating the importance of the $\mathbf{IGAF}$ module.}
			\label{nyuv2ablation1}
			\begin{tabularx}{\textwidth}{CCCC}
				\toprule
				{\bf Fusion Method} & {\bf Addition} & {\bf Concatenation} & \boldmath{$\mathbf{IGAF}$} \\
				\midrule
				$\times 4$ & 1.23 & 1.22 & $\mathbf{1.12}$\\\bottomrule
			\end{tabularx}
			
		\end{table}
		
		
		We also study the effects of different settings of the $\mathbf{IGAF}$ module. The tested settings are (1) skip connections after the $\mathbf{WF}$ modules to propagate deeper features, and not between the $\mathbf{FE}$ and $\mathbf{WF}$ modules, (2) an additional $\mathbf{IGAF}$ module (four in total in the ablation model), (3) the $\mathbf{IGAF}$ module without MLP layers, i.e., the element-wise additions are not weighted, (4) MLP layers consisting of only one dense layer instead of two, and lastly (5)  removing the $\mathbf{WF}$ module. Table \ref{nyuv2ablation2} shows the importance of each component empirically.
		
		\begin{table}[H]
			\caption{Results on the NYU v2 data set.}
			\label{nyuv2ablation2}
			\begin{adjustwidth}{-\extralength}{0cm}
				\begin{tabularx}{\fulllength}{CCCCCCC}
					\toprule
					{\bf Test} & {\bf Relocated Skip} & {\bf Extra} $\mathbf{IGAF}$ & {\bf Without} & {\bf One Layer} & {\bf Without} & {\bf Full Model} \\
					& {\bf Connection} & {\bf Module} & {\bf Weights} & {\bf MLP}&  $\mathbf{WF}$& \\
					\midrule
					$\times 4$ & 1.14 & 1.14 & 1.17 & 1.15 & 1.14 & $\mathbf{1.12}$ \\\bottomrule
				\end{tabularx}
			\end{adjustwidth}
		\end{table}
		

		We note that relocating the skip connection is not a good choice as propagating shallower features of high frequency has more spatial information. The additional module also does not improve the performance as the parameters of the model are increased. This larger model tends to overfit the training data. Keeping the weights of the addition improves the performance as the two parts are dynamically combined after the model has learned which features of each modality are important. The reduction of the MLP layers makes the approximation of the weights weaker which additionally supports the reasoning for using a two-layer MLP combined with our previous ablation. Lastly, without $\mathbf{WF}$, we lack the ability to dynamically increase our feature processing receptive fields that are provided by this module; as such, we lose the ability to capture multi-resolution features effectively. 
		
		\section{Conclusions}
		
		Given the importance of depth perception in its various applications, the ability to estimate accurate, higher resolution depth information is crucial. We proposed an incremental guided attention fusion model for depth super-resolution that uses structural guidance from the RGB modality to provide intermediate structure to the processed features in every layer of the model which leads to the resultant HR output depth map being more accurate compared to the existing methods, as well as free of blurring effects and distortions. Our model's main component, the $\mathbf{IGAF}$ module, performs a cross-modal attention fusion that fuses RGB and depth modalities, while simultaneously focusing on the important information in the intermediate fused features. We achieve state-of-the-art performance on four benchmark datasets against all evaluated baselines for our metrics where the LR depth maps were downsampled from the HR ground truths. Specifically, we demonstrate the ability of our model to generate high-quality depth super-resolutions by training only on the NYU v2 dataset as well as the ability of our model to generalize in a zero-shot setting on the RGB-D-D, Lu and Middlebury datasets in a zero-shot setting which shows the robustness of our method. Additionally, on a $\mathrm{5}$th data set where the LR and HR depth maps were collected using different sensors, mimicking a real-world scenario, we also demonstrate better results than all existing methods.
		
		
		\vspace{6pt}
		\authorcontributions{Conceptualization, A.T., C.K., K.J.M., H.D., D.F. and R.M.-S.; methodology, A.T., K.J.M., H.D. and C.K.; software, A.T. and R.M.-S.; validation, A.T., K.J.M. and C.K.; formal analysis, C.K., K.J.M. and A.T.; investigation, A.T.; resources, C.K., K.J.M., H.D., R.M.-S. and D.F.; data processing, A.T.; writing---Original draft preparation, A.T., C.K. and K.J.M.; writing---Review and editing, A.T., C.K. and K.J.M.; visualization, A.T.; supervision, D.F. and R.M.-S.; project administration, D.F. and R.M.-S.; funding acquisition, D.F. and R.M.-S. All authors have read and agreed to the published version of the manuscript.}
		
		\funding{D.F. acknowledges funding from the Royal Academy of Engineering Chairs in Emerging Technologies programme and the UK Engineering and Physical Sciences Research Council (grant n. EP/T00097X/1).
			R.M.S. and C.K. received funding from EPSRC projects Quantic EP/T00097X/1
			and QUEST EP/T021020/1 and from the DIFAI ERC Advanced Grant proposal 101097708, funded by the UK Horizon guarantee scheme as  EPSRC project EP/Y029178/1. This work was in part supported by a research gift from Google.}
		
		\institutionalreview{The study was conducted in accordance with the Declaration of Helsinki, and approved by the Ethics Committee of the University of Glasgow (application number 300220059 16 December 2022).}
		
		
		
		\dataavailability{The data underlying the results presented in this paper are  available via their open-sourced links cited in the paper.} 
	
	
	\conflictsofinterest{The authors declare no conflicts of interest.} 
	
	\begin{adjustwidth}{-\extralength}{0cm}
 
%

		\reftitle{References}
		

		\PublishersNote{}
	\end{adjustwidth}
\end{document}